\documentclass[twocolumn,10pt]{asme2e}

\usepackage{color}
\usepackage{booktabs}
\usepackage{graphicx}
\usepackage{dblfloatfix}
\usepackage{multirow} 
 \usepackage{array}
 \usepackage{float}
 \usepackage{longtable}
 \usepackage{balance}
 
 \usepackage[table,xcdraw]{xcolor}
\usepackage{appendix}
\usepackage{adjustbox}
\usepackage{amsmath}
\usepackage{flafter} 
\usepackage{multirow}
\usepackage{rotating}
\usepackage{amsmath}
\usepackage{amssymb}
\usepackage{subfigure}
\usepackage{hyperref}
\usepackage[ruled,norelsize]{algorithm2e}
\usepackage[compact]{titlesec}
\titlespacing{\section}{0pt}{5ex}{0.5ex}
\titlespacing{\subsection}{0pt}{3ex}{1ex}
\titlespacing{\subsubsection}{0pt}{1ex}{2ex}

\makeatletter
\newcommand{\removelatexerror}{\let\@latex@error\@gobble}
\makeatother

\confshortname{IDETC/CIE 2021}
\conffullname{the 2021 ASME International Design Engineering Technical Conferences \& Computers and Information in Engineering Conference}

\confdate{17-20}
\confmonth{August}
\confyear{2021}
\confcity{Virtual}
\confcountry{Online}

\papernum{DETC2021-70840}

\title{Classifying Component Function in Product Assemblies with Graph Neural Networks}

\author{Vincenzo Ferrero, Bryony DuPont
	\affiliation{Design Engineering Laboratory\\
	Oregon State University\\
	Corvallis, Oregon, 97331\\
    Email: Ferrerov@oregonstate.edu\\
    Email: Bryony.DuPont@oregonstate.edu\\
   
 }
} 

\author{Kaveh Hassani, Daniele Grandi
	\affiliation{Autodesk, Inc.\\
	Autodesk Research\\
	San Rafael, CA 94903\\
 Email: Kaveh.Hassani@autodesk.com \\
 Email: Daniele.Grandi@autodesk.com 
 }
}

\begin{document}

\maketitle    

\begin{abstract} 
{\it Function is defined as the ensemble of tasks that enable the product to complete the designed purpose. Functional tools, such as functional modeling, offer decision guidance in the early phase of product design, where explicit design decisions are yet to be made. Function-based design data is often sparse and grounded in individual interpretation. As such, function-based design tools can benefit from automatic function classification to increase data fidelity and provide function representation models that enable function-based intelligent design agents. Function-based design data is commonly stored in manually generated design repositories. These design repositories are a collection of expert knowledge and interpretations of function in product design bounded by function-flow and component taxonomies. In this work, we represent a structured taxonomy-based design repository as assembly-flow graphs, then leverage a graph neural network (GNN) model to perform automatic function classification. We support automated function classification by learning from repository data to establish the ground truth of component function assignment. Experimental results show that our GNN model achieves a micro-average F${_1}$-score of 0.832 for tier 1 (broad), 0.756 for tier 2, and 0.783 for tier 3 (specific) functions. Given the imbalance of data features, the results are encouraging. Our efforts in this paper can be a starting point for more sophisticated applications in knowledge-based CAD systems and Design-for-X consideration in function-based design.

}
\end{abstract}



\section{INTRODUCTION}
Function-based design is a foundational tenet in product design \cite{Ullman2003}. \textit{Function} is defined as the application of the product purpose toward solving a design problem \cite{Gero2004,Rosenman1998}. Components within the product complete sub-functions necessary to materialize the overarching product function. In product design, functional modeling is used to support and guide designers during early conceptual design phases \cite{Eisenbart2011,Eisenbart2013}. Here, a designer determines the sub-functions needed to complete the primary product function and purpose \cite{Ullman2003}. These sub-functions are connected through flows that capture their interactions. In practice, these flows represent material, energy, and signal transfer \cite{Hirtz2002}.

Currently, function-based design suffers from subjectivity caused by the designer's interpretation of function and flow as it applies to a design. Efforts have been made to standardize function and flow into taxonomies to limit subjectivity, while increasing shared domain understanding \cite{Hirtz2002}. The standardization of function-based design principles has led to meaningful curation of taxonomy-based design repositories \cite{Ferrero2018,Oman2014,Szykman1999}. While these design repositories have been widely accepted into literature, there remain challenges in function interpretation defined by designer expertise in function-based design. The human interpretation and assignment of function have generated repositories that are often unorganized, sparse, and unbalanced.

Low data quality and scarcity of design repositories have led to an under-utilization of deep learning methods in the data-driven product design field \cite{feng2020,bertoni2020}, as they require large amounts of data \cite{halevy_2019, sun_2017}. Prior work addressed the issue of scarce structured design knowledge datasets by automatically extracting function knowledge from a corpus of mechanical engineering text to construct a design knowledge base \cite{Cheong2017}. For modalities of data other than text, researchers relied on synthetic design data \cite{law2020}, small amounts of curated knowledge \cite{zhang2019cad}, or scraped public online design repositories and manually labeled design knowledge \cite{angrish2019, dering2017}. Yet, it remains challenging to apply data-driven methods in the field of mechanical design \cite{han2020semantic, lupinetti2019}. However, recent progress in graph representation learning and graph neural networks (\textbf{GNNs}), show promise in knowledge discovery in sparse datasets \cite{zhang_2020_kde, wu_2020_nnls}. Rapid advancements in deep learning for sparse datasets present an opportunity to apply such methods on design repository data to forward the state-of-the-art in data-driven design, specifically in the context of function-based design shared understanding, standardization, and computer representation.

In this paper, we use GNNs and sparse data from a design repository to classify component function based on assembly and flow relationships. We represent data from a hierarchical taxonomy-based design repository through graphs. The focus of these graphs is to capture function-flow-assembly relationships within products housed in the design repository. We then introduce a hierarchical GNN framework that capitalizes on the three-tier hierarchical nature of the repository data. Using the hierarchical GNN, we classify component function in three tiers ranging from broad primary functions to detailed tertiary functions as introduced in previous literature \cite{Hirtz2002}. We exhaustively evaluate our GNN framework using four types of GNN layers and compare its results against other feed-forward networks to determine the fidelity of our proposed GNN architecture. We also compare our hierarchical GNN architecture against independently trained GNNs for each component function tier. The performance of our GNN framework is presented and subsequently explored through confusion matrices and feature importance analysis.

\subsection{Specific Contributions}

The research presented here contributes to the area of function-based data-driven product design by leveraging recent developments in graph representation learning to enable a more descriptive shared understanding between humans and computers about the function of parts in an assembly. Our interest in function classification stems from recent work that applies data-driven approaches to various engineering design tasks \cite{feng2020}, such as searching a design space \cite{bang2018}, model-based systems engineering \cite{berquand2019}, or selecting appropriate manufacturing methods \cite{angrish2019}. Such work points towards intelligent design agents enabled by knowledge-based design systems, which have been explored by the design research community over many years \cite{coyne1990,feng2020}. 

\footnotetext[1]{\url{https://github.com/VincenzoFerrero/OSDR-GNN}}

In the context of our work, functional modeling supports the use of automated reasoning systems, as well as facilitating communication and understanding between designers and co-creative agents, both of which could benefit from a better-shared understanding of the problem when working on a creative task \cite{erden2008,davis2015,law2020}. We see function as an important theoretical element to allow an intelligent design agent to better understand the designer's intent when co-creating with the designer. Predicting low-level functions of a design is an initial step towards this vision. The work we present here contributes the following:
\begin{enumerate}
    \item A novel approach to automatically predict the function of a part in an assembly using graph neural networks.
    \item A publicly available relational assembly graph model to represent design repository data.
    \item Experimental results of part function classification from a graph representation of the assembly.
\end{enumerate}
In this body of work, we use the Oregon State Design Repository (\textbf{OSDR}) as our structured taxonomy-based data source \cite{Bohm2004,Bohm2008}. We provide a publicly available subset of the OSDR dataset used in our work, the assembly graphs representing the OSDR data, and the GNN implementation for the research community to leverage in future work \footnotemark[1]. Furthermore, graph representation of the OSDR can be leveraged in data searching tasks and other GNN tasks beyond function classification. GNN and OSDR graph representations can be used to ascertain design knowledge on material choice, assembly order, failure modes, \textit{Design for the Human Element}, \textit{Design for the Environment} and component-system classification tasks.

\section{BACKGROUND} 
In this section, we introduce fundamental concepts and research supporting the extraction of functional knowledge using GNNs. Here, we introduce literature in function-based product design in the context of design support and deep learning. Next, design repositories are discussed as a source of semantic product data that can be used in modern deep learning techniques. Finally, literature and background are established for graph representation and GNNs.

\subsection{Function-based Product Design}
Function-based design has been used as a bridge to bring Design-for-X (DfX) objectives, such as Design for the Environment, from post-design analysis to the earlier design phases of product development. To this end, function-based design has been used with life-cycle assessment data to provide function-based sustainable design knowledge to designers \cite{Arlitt2017,Devanathan2010,Gilchrist2012}. In human-centered product design, function has been related to human error and interaction points to determine which functions need special consideration for ergonomics \cite{SoriaZurita2020,SoriaZurita2018}. These recent developments in function-based design for meeting DfX objectives suggest a need to predict, learn from, and model function in components as a means to bring further curated data to early design phases.

Previous efforts have been made to use machine learning for improving function-based design methodologies. In other research, association rules and weighted confidence has been used to determine the function of a component within product configurations \cite{Tensa2019,Mikes2020a,Edmonds2020}. Decision trees have proved useful in reducing the feasible design space of functional assignment when considering product assembly \cite{Ferrero2020}. Furthermore, deep learning approaches have been used to disambiguate customer reviews based on function, form, and behavior \cite{Singh2017}.

\subsection{Taxonomy-based Design Repositories and Knowledge Discovery}
Taxonomy-based design repositories store product design data relevant to design engineers \cite{Szykman2000,Bohm2004}. This type of design repository is generated through expert taxonomy descriptions of classical product life cycle inventory (LCI) data. For example, given common product data such as a bill of materials, specialized taxonomy data can be appended to LCI data. The OSDR is a taxonomy-based repository that houses product LCI data along with assigned specialized taxonomy descriptions \cite{Ferrero2018,Bohm2008}. In the OSDR case, specialized taxonomy data includes product assembly child-parent notation, functional-flow basis assignment to components, and a standardized component naming schema.

Adoption of design repositories in research and industry has been slow due to resource commitment, human curation, intuition-based knowledge extraction, and lack of well-structured product data. Efforts have been made to improve design repository generation by limiting subjectivity through taxonomy standardization \cite{Phelan2014,Bharadwaj2019,Kurtoglu2005,Hirtz2002,Cheong2011}. Furthermore, recent approaches have been introduced to streamline data addition to design repositories \cite{Ferrero2020a}. Despite the described challenges, design repositories have been shown to be useful in data-driven design approaches, particularly in machine learning and knowledge extraction tasks. 

Design repositories are useful in knowledge discovery tasks and have been effectively employed within machine learning approaches
\cite{Fayyad1996,Fayyad1996a,Fayyad1996b,Williams2019}. Specifically related to function-based design, design repositories were used in the automated extraction of function knowledge from text \cite{Cheong2017}. In our work, we assert that recent advancements in graph representation learning have allowed for the ability to generate predictive models from sparse, incomplete, subjective, and otherwise unbalanced repository data. 

\subsection{Graphs in Product Design}
Graphs are robust data structures that represent interactions (i.e., edges) among constituents (i.e., nodes) of a system. They can also capture the direction of interactions, properties of interactions (i.e., edge attributes), and properties of the system constituents (i.e., node attributes). In product design, knowledge graphs \cite{wang2017knowledge, ji2020survey} which are a specific type of graph that represent structural relations between entities of a domain, are widely used. Classically they are most often used in natural language processing tasks \cite{Miller1995,Liu2004,Sarica2020}. Current efforts in knowledge graphs have facilitated robust graph representation of domain-specific semantic relationships of product design \cite{Sarica2019}. TechNet was developed in 2019 by mining semantic relationships of elemental concepts found in US patent data. B-link was introduced in 2017 by mining engineering domain knowledge from engineering-focused academic literature \cite{Shi2017}. Knowledge graphs have supported product design by providing language and design relationships. Specifically, engineering design knowledge graphs have been used in concept generation and evaluation \cite{Han2020}. However, there is a need to expand on knowledge graphs with standardized product design resources, such as design repositories. 

Recently, a knowledge graph framework has been introduced to create rich node and edge features based upon taxonomy-based product design models \cite{Zhang2019}. Here, graphs generated with product design taxonomies capture meaningful relationships between product materials, manufacturing method, tolerance, function, and other product features. Specifically, product design knowledge graphs have been useful for case-based reasoning and concept similarity search. In this work, we expand on taxonomy-based graphs with the representation of repository data in a graph structure, with a focus on function, flow, and assembly representations. The generated graphs are then used in prediction tasks using GNNs.

\subsection{Graph Neural Networks}
Standard deep learning architectures such as convolutional neural networks (CNN) and recurrent neural networks (RNN) operate on regular-structured inputs such as grids (e.g., images, volumetric data) and sequences (e.g., signals, text). Nevertheless, many real-world applications deal with irregular data structures. For instance, molecular structures, interaction among sub-atomic particles, or robotic configurations cannot be reduced to a sequence or grid representation. Such data can be represented as graphs, which allow for jointly modeling constituents of a system, their properties, and interactions among them. GNNs \cite{hassani2020contrastive, li2015gated, hamilton_2017_nips, kipf_2017_iclr, velickovic_2018_iclr, xu_2019_iclr, duvenaud_2015_nips} can directly take in data structured as graphs and use the graph connectivity as well as node and edge features to learn a representation vector for every node in the graph. Because GNNs utilize the strong inductive bias of connectivity information, they are more data-efficient compared to other deep architectures. GNNs have been successfully applied to point clouds and meshes \cite{hanocka2019meshcnn, hassani_2019_iccv}, robot designs \cite{wang_2018_iclr}, physical simulations\cite{sanchez_2018_icml, sanchez2020learning}, particle physics \cite{shlomi2020graph}, material design \cite{guo2020semi}, power estimation \cite{park2019physics}, and molecule classification \cite{duvenaud_2015_nips}.

Let $G = (V, E)$ denote a graph with vertices (node) $V$, edges $E$, node attributes $X_v$ for $v \in V$ and edge attributes $e_{uv}$ for $(u, v) \in E$. Given a set of graphs $\left\{G_1, . . . , G_N \right\}$ and their node labels $\left\{y_{v_1}^1, . . . ,y_{v_m}^1, . . . ,y_{v_1}^N, . . . , y_{v_k}^N\right\}$, the task of supervised node classification is to learn a representation vector (i.e., embedding) $h_v$ for every node $v\in G$ that helps predict its label. GNNs use a neighborhood aggregation approach, where representation of node $v$ is iteratively updated by aggregating representations of neighboring nodes and edges. After $k$ iterations of aggregation, the representation captures the structural information within its $k$-hop network neighborhood \cite{gilmer_2017_icml}. Formally, the $k$-th layer of a GNN is defines as:
\begin{equation}
h_v^{(k)} = f_\theta^{(k)}\left(h_v^{(k-1)}, g_\phi^{(k)}\left(\left\{h_v^{(k-1)}, h_u^{(k-1)}, e_{uv} : u \in N(v) \right\}\right.\right)
\end{equation} 
where $h_v^{(k)}$ is the representation of node $v$ at the $k$-th layer, $e_{uv}$ is the edge feature between nodes $u$ and $v$, and $N(v)$ denote neighbors of $v$. $f_\theta(.)$ denotes a parametric 
combination function and $g_\phi(.)$ denotes an aggregation function. We initialize $h_v^{(0)}= X_v$.

Different instantiations of $f_\theta(.)$ and $g_\phi(.)$  functions result in different variants of GNNs. In this paper, we compare the performance of four well-known variants of GNNs including GraphSAGE \cite{hamilton_2017_nips}, graph convolution network (GCN) \cite{kipf_2017_iclr}, graph attention network (GAT) \cite{velickovic_2018_iclr}, and graph isomorphism network (GIN) \cite{xu_2019_iclr}. For an overview of GNNs see \cite{zhang_2020_kde, wu_2020_nnls}.

\section{METHODS}
In this section, we describe the methodology for function classification using a GNN on repository data that is represented by graphs. First, the data selection and processing is presented.  Then, we describe graph schema and graph construction. Finally, a GNN and related parameters are introduced to predict the hierarchical functions. 

\subsection{Data Selection and Processing}
\subsubsection{Data Selection}
The OSDR is a function-based relational framework built upon consumer product component (artifact) information \cite{Bohm2008,Bohm2004,Szykman1999}. The repository schema includes system-level bill of materials, system type, component function and flow, material, and assembly relationships.
The data within the OSDR utilizes published standard taxonomies for component, function, and flow naming. These standard taxonomies are referred to as basis terms \cite{Kurtoglu2005,Hirtz2002}. The basis taxonomies feature a hierarchy system that allows for broad-to-specific identification of component name, function, and flow per component artifact in the OSDR.  In the taxonomy hierarchy systems within the OSDR, tier 1 basis terms encompass the broadest description of the basis term. In ascending order, tier 2 and 3 increase specificity, information, and accuracy of the basis definition. An example of the basis term hierarchy from each taxonomy is shown in Table \ref{table:basis}. This example is not representative of any product within the OSDR and only demonstrates the hierarchical structure of taxonomy basis terms.

\begin{table}[h]
  \centering
  \caption{\uppercase{ \footnotesize Example Hierarchy for Component (Supporter), Function (Branch), and Flow (Signal) basis terms}}
  \begin{small}
  \setlength\extrarowheight{-4pt}
  \begin{tabular}{@{}m{2.5cm}m{2.9cm}m{1.5cm}@{}}
    \toprule
    \textbf{Primary (Tier 1)} &\textbf{Secondary (Tier 2)} & \textbf{Tertiary  (Tier 3)} \\
    \midrule
    \midrule
    & \textbf{Component} &  \\
    \midrule
    Supporter & Stabilizer & Insert \\
              &            & Support \\
              & Positioner & Washer \\
              &            & Handle \\
              & Securer & Bracket \\
    \midrule
    \midrule
    & \textbf{Function} &  \\
    \midrule
    Branch & Separate & Divide \\
           &          & Extract \\
           &          & Remove \\
           & Distribute &  \\
    \midrule
    \midrule
    & \textbf{Flow} &  \\
    \midrule
    Signal & Status & Tactile \\
           &          & Taste \\
           &          & Visual \\
           & Control & Analog  \\
           &         & Discrete  \\
    \bottomrule
  \end{tabular}
  \end{small}
  \label{table:basis}
\end{table}
 The OSDR encapsulates the data of 184 consumer products and 7,275 related artifacts. Artifacts are generally components but can also represent sub-assemblies and systems. Artifacts are related through parent-child a familial hierarchy (hypernym and hyponym relations). Functional relationship and product-level functional models are captured through component-level function, input flow, and output flow. In this regard, the OSDR houses 19,627 component-related function data points with 19,667 corresponding flow data points.

\subsubsection{Processing}
For our methodology, the data from the OSDR needed to be filtered and processed prior to developing the product graphs. We removed 24 consumer products from the dataset due to a lack of completion in function, flow, or assembly definition. From the 160 products, data points are represented by a single component defined by material, component basis, parent component, functional basis hierarchy, input flow, output flow, input component, and output component. Each data point is a unique representation of the components defined by flow attributes. Concisely, there are many data points per component depending on the number of functions and related flows managed by that component. The processing and filtering of the data within the OSDR resulted in 15,636 data points represented by 137 component basis terms, 51 function basis terms, 36 flow basis terms, and 16 material categories. An example vegetable peeler product with component, function and flow data is shown in Table \ref{BOM} found in appendix \ref{appendix:data}.

\subsection{Assembly-Flow Graph Generation}
\label{KG}
\begin{figure}[h!]
    \centering
    \includegraphics[width=6cm]{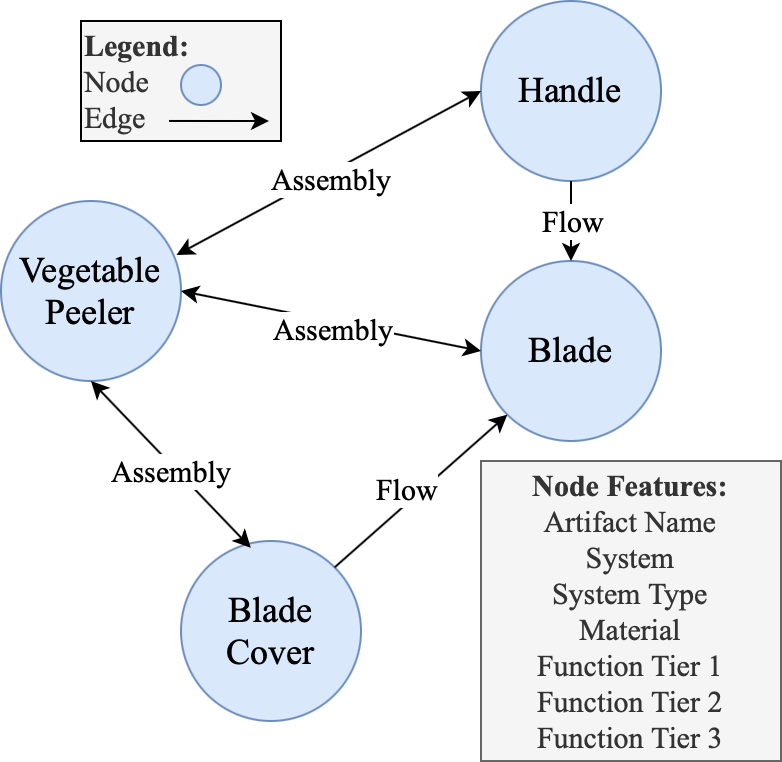}
    \caption{\uppercase{\footnotesize simplified relational assembly graph  example of components from Table \ref{BOM}}}
    \label{fig:GraphSchema}
\end{figure}

The processed OSDR data is represented through relational graphs. Relational graphs are a specific type of graphs with the following properties: (1) They are directed graphs, meaning edges between nodes have directions, (2) they are also attributed, meaning that the graphs contain node and edge attributes, and (3) they are multi-graphs, as more than one edge is allowed between any two nodes.\. Graph representation of the OSDR is needed to apply our proposed GNN architecture outlined in section \ref{GNN}. In the context of the OSDR, the relational graphs are generated per system to represent the assembly and flow relations within the system. These assembly-flow graphs are defined by nodes and connecting edges. Figure \ref{fig:GraphSchema} shows the graph structure. The graphs are generated using predefined schema definitions from the OSDR. These definitions can be found in related literature and explored through the hosted version of the OSDR \footnotemark[2] \cite{OSDR}.

\footnotetext[2]{\url{http://ftest.mime.oregonstate.edu/repo/browse/}}
The nodes are representative of each artifact data point and carry the following features: system name, system type, component basis term, material, and functional basis hierarchy. The nodes are connected through two edge types denoted as flow edges and assembly edges.

\begin{figure*}[t]
    \centering
    \includegraphics[width=170mm]{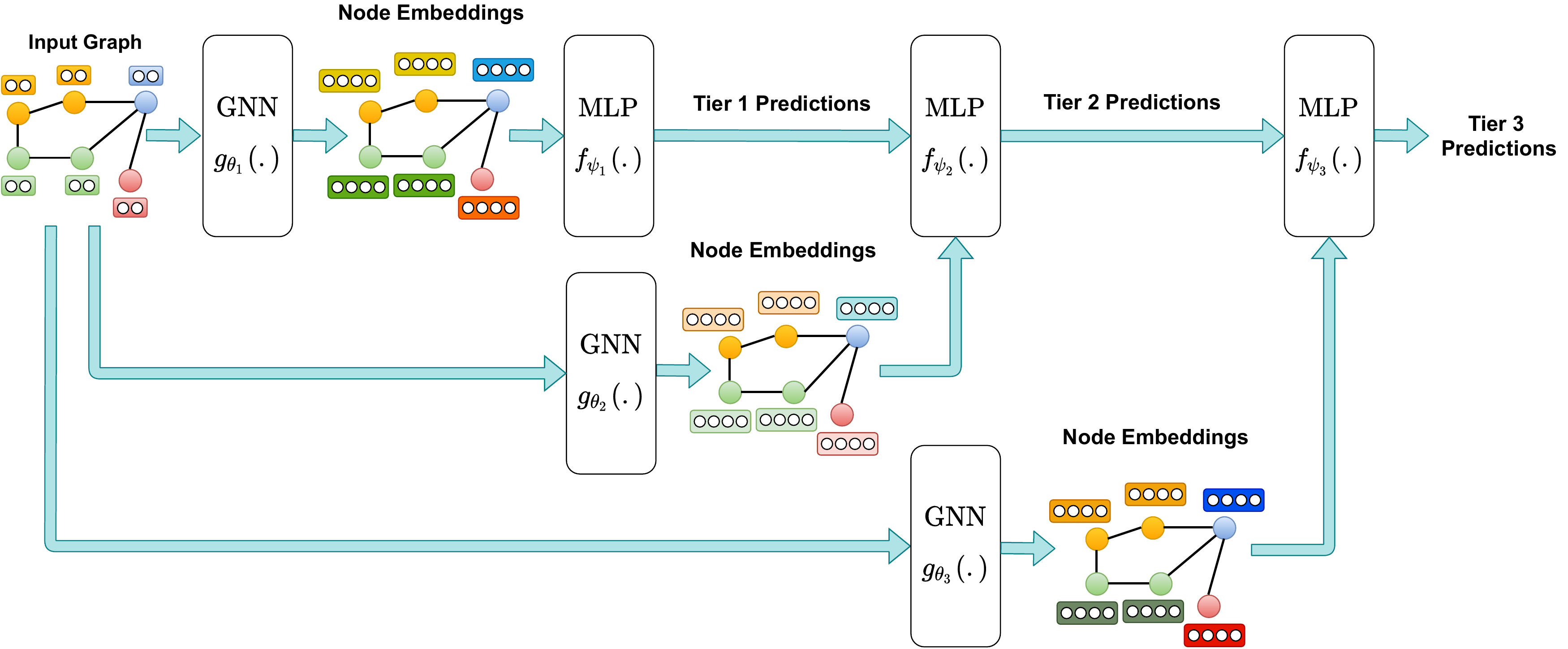}
    \caption{\uppercase{\footnotesize The proposed hierarchical Graph Neural Network Framework}} 
    \label{gnn}
\end{figure*}

A \textit{flow edge} is defined as the connections between functions within the system and the movement of energy, signals, and materials through a system in which a function or set of functions modifies. A function has an input flow and an output flow. Flow edges are directional and defined by the data specified flow basis representation regardless of flow basis hierarchy. In this regard, we are not backing out higher-tier flow labels. Flow edges are defined by only the assigning flow basis term found within the data.  Flow edges are only represented once per input-output relationship. Assembly edges are non-directional physical connections between artifacts in the classical product assembly sense. Both assembly and flow edges are used to capture the totality of physical-functional interaction between artifacts. By representing both physical connections (assembly) and function connections (flow) through edge definitions, we aim to classify component function using late-stage design defined physical product assembly and early design stage defined function-flow definition. Considering the totality of the product design process increases the breadth of contribution of this work.

\subsubsection{Dataset Metrics}
We generated 160 assembly graphs representative of the 160 non-filtered products from the OSDR. NetworkX \cite{SciPyProceedings_11}, a Python library for graph processing, is used to materialize the relational graphs \cite{hagberg2008exploring}. Per graph, there is an average of 98 nodes and 791 edges. When singling out flow edge type, there is an average of 537 flow edges per graph. For assembly edges, there is an average of 262 edges per node.

\subsubsection{Assembly-Flow Graph Processing}
\begin{figure}[h!]
    \centering
    \subfigure[]{\includegraphics[width=.39\textwidth]{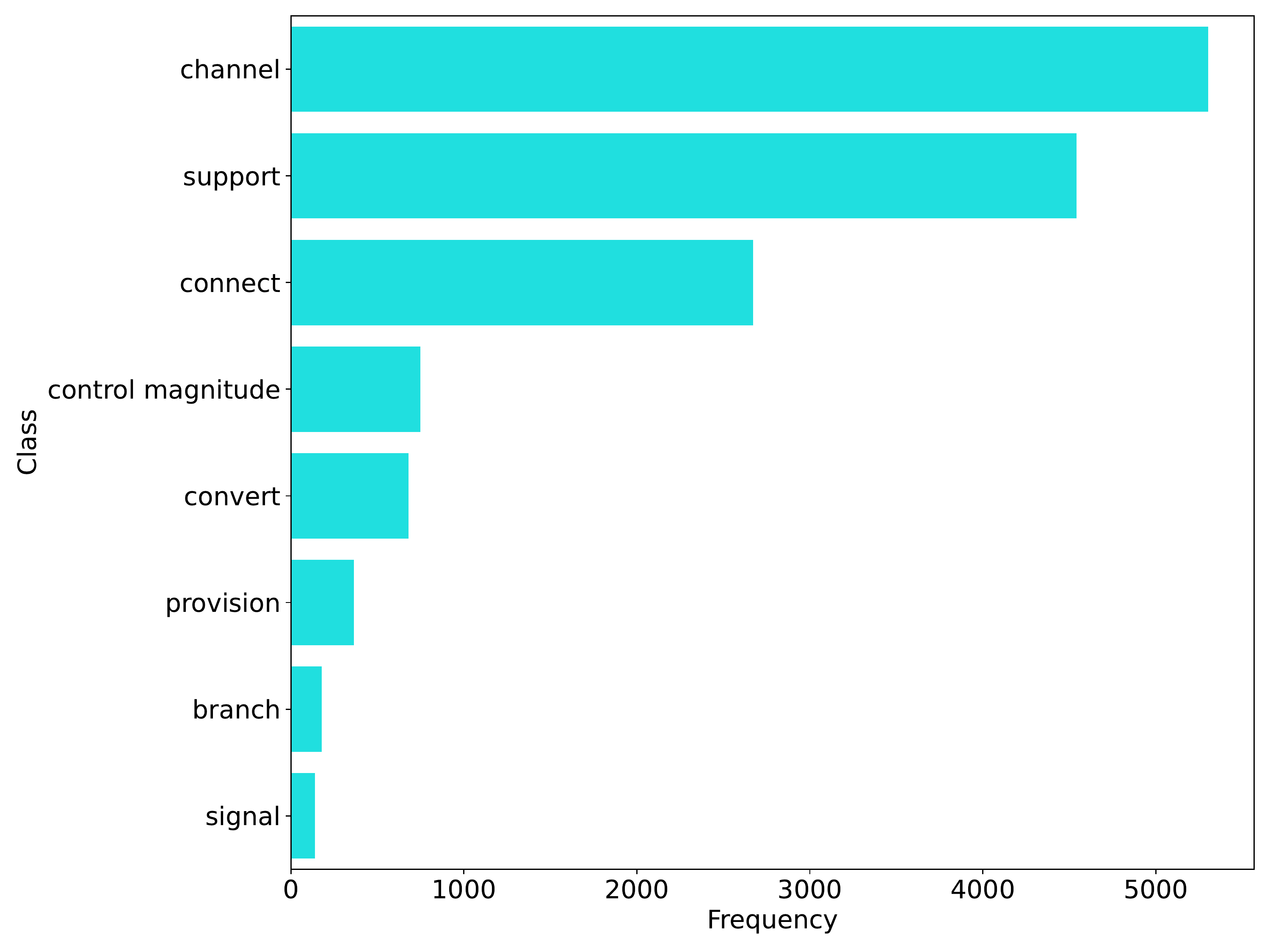}} 
    \subfigure[]{\includegraphics[width=.39\textwidth]{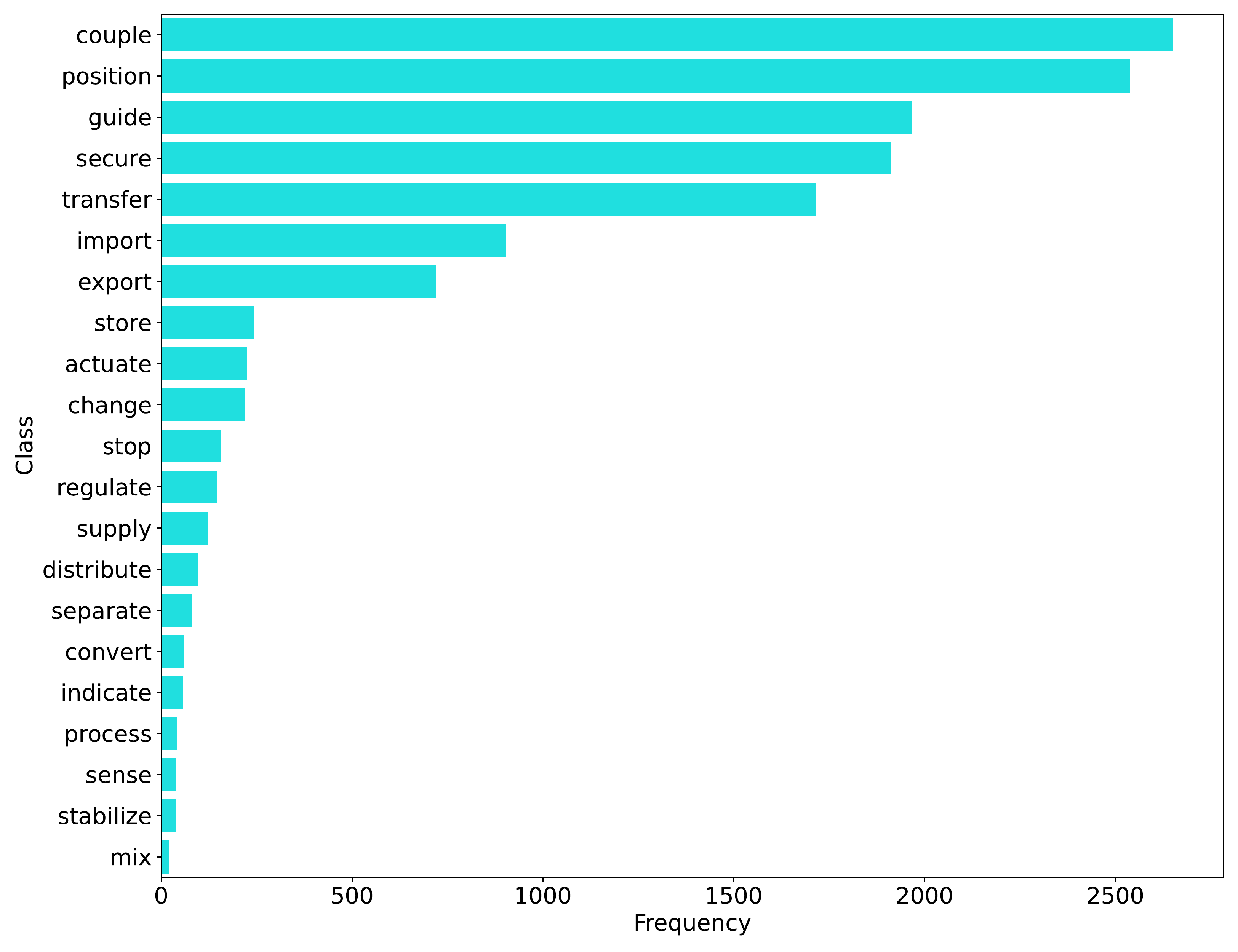}} 
    \subfigure[]{\includegraphics[width=.39\textwidth]{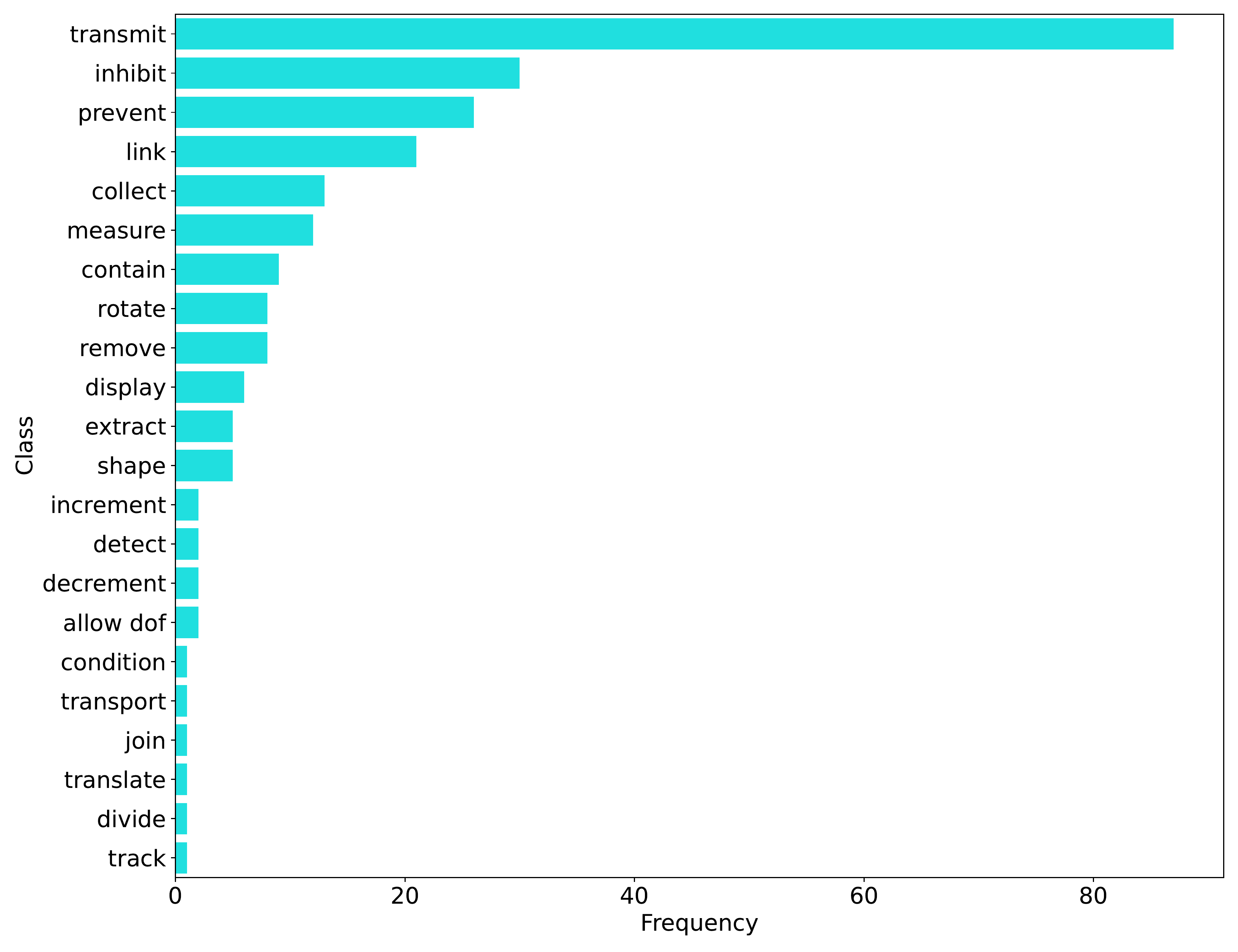}}
    \caption{\uppercase{Distribution of class frequencies in (a) tier 1 (b) tier 2 (c) tier 3 function categories.}}
    \label{fig:frequency}
\end{figure}

We pre-process the data as follows. For initial node attributes, we concatenate one-hot encoding of component basis, system name, system type, and material features resulting in a 316-dimensional multi-hot initial node feature. For edge attributes, we concatenate one-hot encoding of input flow, output flow, and an indicator of whether the edge represents an assembly connection. This results in a 75-dimensional initial edge feature. The dataset contains 9, 22, and 23 category labels for tiers 1, 2, and 3 functions, respectively.  It is also noteworthy that label distribution in all three tiers is highly skewed. The label frequencies are shown in Figure \ref{fig:frequency}.

\subsection{Learning Architecture}
\label{GNN}

Inspired by recent advances in graph representation learning, our approach learns dedicated node representations for each functional tier prediction task. As shown in Figure \ref{gnn}, our method consists of three GNN encoders that take in graphs connectivity information along with initial node and edge features and produce dedicated node embeddings for each tier. Each GNN is then followed by a dedicated multilayer perceptron (MLP) that acts as a specialized classifier for that tier. Furthermore, we utilize the hierarchical nature of function tiers to augment the predictions and use hierarchically structured local classifiers with a local classifier per tier. 

Assume a training set $\mathcal{D}=[G_1, G_2,...,G_N]$ of  $N$ graphs where each graph is represented as $G=(\textbf{A}, \textbf{X}, \textbf{E}) $ where $\textbf{A}\in\lbrace 0, 1 \rbrace^ {n\times n}$ denotes the adjacency matrix 
(one-hop connectivity information), $\textbf{X}\in \mathbb{R}^{n \times d_x}$ is the initial node features, and $\textbf{E}\in \mathbb{R}^{n \times n \times d_e}$ is the initial edge features. We define three GNNs 
$g_{\theta_k}(.): \mathbb{R}^{n \times n}  \times  \mathbb{R}^{n \times d_x} \times \mathbb{R}^{n \times n \times d_e} \longmapsto \mathbb{R}^{n \times d_h}, k=\{1,2,3\}$ parametrized by $\left\{\theta_k\right\}_{k=1}^3$ corresponding to tiers 1 to 3 functions, respectively. This results in three sets of dedicated node embeddings $\textbf{H}^{t_1}, \textbf{H}^{t_2}, \textbf{H}^{t_3} \in \mathbb{R}^{n \times d_h}$. GNNs are essentially learning to extract strong representations for down-stream classifiers. We use three MLPs $f_{\psi_k}(.): \mathbb{R}^{n \times (d_h + |Y_{k-1}|)} \longmapsto \mathbb{R}^{n \times |Y_k|}, k=\{1,2,3\}$ parameterized by $\left\{\psi_k\right\}_{k=1}^3$ where $k$-th MLP is the dedicated classifier for predicting tier $k$ function classes. The $k$-th MLP receives the learned node representations from its dedicated GNN $g_{\theta_k}(.)$ (i.e., $\textbf{H}^{t_k}$) and predictions of the predecessor MLP in hierarchy $f_{\psi_{k-1}}(.)$ to predict the function classes for $k$-th tier. Because the first MLP does not have any predecessors (i.e., first tier in hierarchy), we simply pass a vector of zeros to emulate the input predictions.

During the training phase, we utilize \emph{teacher forcing} \cite{williams1989learning} to enhance the training process. Teacher forcing is a procedure in which during training, the model receives the ground truth output (rather than predicted output) as input at the next step. In other words, rather than feeding the $k$-th MLP with the actual predictions of $(k-1)$-th MLP, we feed it with ground truth labels of $(k-1)$-th tier labels. During inference, however, we do not have access to the ground truth labels. Therefore, we feed the subsequent MLP with the probability distribution of the predicted labels. We use Softmax function over the MLP predictions to transfer the raw predictions into proper probability distributions. Furthermore, we use frequency-based weighting to address the data imbalance during training. We compute the loss such that less frequent classes contribute more to the total loss compared to frequent classes. This practice prevents the model from paying more attention to frequent classes and ignoring the rare ones. We jointly optimize the model parameters with respect to the aggregated and weighted cross-entropy losses of all three functional tier predictions using mini-batch stochastic gradient descent. The process of training for one mini-batch is shown in Algorithm \ref{algo}.
\begin{figure}[h!]
    \removelatexerror
    \begin{algorithm}[H]
    \SetAlgoLined \DontPrintSemicolon
    \caption{Training the proposed model with teacher forcing for one mini-batch. $\textbf{H}^{t_k}, Y^{t_k}_p, Y^{t_k}_g$ denote learned node representation, predicted labels, and ground 
    truth labels for tier $k$ function hierarchy.}\label{algo}
   \KwIn{Cross-entropy loss $\mathcal{L}$, GNNs $g_{\theta_k}(.)$, MLPs $f_{\psi_k}(.)$, sampled batch of N graphs $\{G_j \}_{j=1}^{N}$, concatenation operator $\parallel$ }\\
$\textbf{L} \gets \emptyset$ \\
\For{$G$ in a batch $\{G_j \}_{j=1}^{N}$}
{
\footnotesize\tcp{Compute dedicated node embeddings}\normalsize \\
$\textbf{H}^{t_1} \gets g_{\theta_1}\left(G\right)$ \\
 $\textbf{H}^{t_2} \gets g_{\theta_2}\left(G\right)$ \\
$\textbf{H}^{t_3} \gets g_{\theta_3}\left(G\right)$ \\
\footnotesize\tcp{Compute tier predictions}\normalsize
$Y^{t_1}_p\gets \text{Softmax}\left(f_{\psi_1}\left(\left[\textbf{H}^{t_1} \parallel \textbf{0}\right]\right)\right)$ \\
$Y^{t_2}_p\gets \text{Softmax}\left(f_{\psi_2}\left(\left[\textbf{H}^{t_2} \parallel Y^{t_1}_g \right]\right)\right)$ \\
$Y^{t_3}_p\gets \text{Softmax}\left(f_{\psi_3}\left(\left[\textbf{H}^{t_3} \parallel Y^{t_2}_g \right]\right)\right)$ \\
\footnotesize\tcp{Compute joint loss across all tiers}\normalsize
$\textbf{L} \gets \textbf{L} + \mathcal{L}\left(Y^{t_1}_p, Y_g^{t_1} \right) + \mathcal{L}\left(Y^{t_2}_p, Y_g^{t_2} \right)  + \mathcal{L}\left(Y^{t_3}_p, Y_g^{t_3} \right)$ \\
}
\footnotesize\tcp{Compute gradients and update parameters}\normalsize
$\left\{\theta_k, \psi_k \right\}_{k=1}^{3} \gets \left\{\theta_k, \psi_k \right\}_{k=1}^{3} - \gamma\nabla_{\left\{\theta_k, \psi_k \right\}_{k=1}^{3}} \frac{1}{N} \sum\limits_{j=1}^N{\textbf{L}} $
\end{algorithm}
\end{figure}

\section{Results and Method Validation}
\label{validation}
In this section, we introduce the GNN architecture implementation and results. We explore results further with confusion matrices to determine function-specific performance. We then validate the results of the SAGE graph neural network algorithm against three other state-of-the-art GNNs. The GNN types are GCN, GAT, and GIN \cite{kipf_2017_iclr,velickovic_2018_iclr,xu_2019_iclr}. In closing, we highlight feature importance to determine the most consequential taxonomy-based data features toward classifying component function and look to investigate how our proposed hierarchical GNN architecture compares to a group of independent GNNs. 

\subsection{Experimental Protocol}
Given the small size of the dataset, we split it into 60\%, 10\%, and 30\% train, validation, and test sets, respectively, using a random distribution, and report the mean and standard deviation of the metrics after running the experiments 100 times. In each run, we split the dataset, train a model on the train set, tune the hyperparameters on the validation set, and report the results on the test set. This allows us to investigate the model's performance without bias towards train/test splits. Also, given the imbalanced nature of the labels in all three functional tiers, we use precision (P), recall (R), and F$_1$-score metrics to report the results. These metrics are defined as follows:
\begin{equation}
\text{P}=\frac{\text{TP}}{\text{TP}+\text{FP}}, \quad \text{R}=\frac{\text{TP}}{\text{TP}+\text{FN}}, \quad \text{F}_{1}=2 \times \frac{\text{P} \times \text{R}}{\text{P} + \text{R}}
\end{equation}
where TP, FP, and FN denote the number of true positive, false positive, and false-negative predictions. Moreover, we report the metrics with three types of averaging: micro, macro, and weighted averaging. Micro-averaging computes F$_1$-score by considering the total number of TP, FP, and FN, whereas macro-averaging computes F$_1$-score for each label and averages it without considering the frequency for each label. On the other hand, weighted-averaging computes F$_1$-score for each label and returns the weighted average based on the frequency of each label in the dataset. In practice, micro-averaging is useful in heavily imbalanced datasets, macro-averaging is useful in balanced datasets, and weighted-averaging is useful in datasets where some features are balanced, and some are not. On all accounts, we are trying to maximize the precision (P), recall (R), and F$_1$-score metrics for each function tier prediction. However, we are most interested in micro-average performance as our dataset is unbalanced and sparse.

We initialize the network parameters using Xavier initialization \cite{glorot_2010_aistat} and train the model using Adam optimizer \cite{kingma_2014_iclr} with an initial learning rate of 1e-3. We use a cosine scheduler \cite{loshchilov2016sgdr} to schedule the learning rate, and also use early-stopping with a patience of 50. We also apply Leakey rectified linear unit (ReLU) non-linearity \cite{maas2013rectifier} with negative slope of 0.2, and dropout \cite{srivastava_2014_jmlr} with probability of 0.1 after each GNN layer. We choose the number of GNN layers and hidden dimension size from the range of [1, 2, 3] and [64, 128, 256], respectively. Finally, we choose the GNN layer type from GraphSAGE \cite{hamilton_2017_nips}, GCN \cite{kipf_2017_iclr}, GAT \cite{velickovic_2018_iclr}, and GIN \cite{xu_2019_iclr} layers. We implemented the experiments using PyTorch \cite{NEURIPS2019_9015} and used Pytorch Geometric \cite{Fey_2019} to implement the GNNs. The experiments are run on a single RTX 6000 GPU where on average, one epoch of training takes about 1 second and 1.5GBs of GPU memory. 

\subsection{Results}
\label{results}
To investigate the performance of the GNNs on the dataset and compare it with other feed-forward networks, we trained an MLP, a logistic regression (linear) model, and four types of GNNs including GraphSAGE \cite{hamilton_2017_nips}, GCN \cite{kipf_2017_iclr}, GAT \cite{velickovic_2018_iclr}, and GIN \cite{xu_2019_iclr}. To train the GNNs, we used the connectivity information along with initial node and edge features, whereas for the MLP and linear models, we only used initial features. The results are shown in Table \ref{table:result}. We observe a classification weighted precision of 0.846  for tier 1 functions, 0.777 for tier 2 functions, and 0.860 for tier 3 functions. When strongly considering the data imbalance (micro-average), we observe a precision of 0.832 for tier 1 functions, 0.757 for tier 2 functions, and 0.787 for tier 3 functions. If we ignore data imbalance (macro-average), there is a precision of 0.882 for tier 1 functions, 0.803 for tier 2 functions, and 0.831 for tier 3 functions. Moreover, results suggest that: (1) GNNs significantly outperform MLP and Linear models in all tiers across all metrics. For example, the best performing GNN in tier 1 function prediction outperforms the MLP model with an absolute F$_1$-score of 0.295, i.e., a relative improvement of 55.24\%. This implies that connectivity information plays an important role in the predictions. (2) Among GNNs, a GNN with GraphSAGE layers slightly performs better in tier 1 function predictions, whereas for tier 2 and
3 predictions, GNNs with GIN and GCN layers perform better, respectively. This shows the importance of treating the GNN layers as hyper-parameters which can yield better performance. 

\subsubsection{Function-Specific Performance}
We also investigate the performance of models on individual labels using confusion matrices. Figures \ref{fig:confusion} shows the confusion matrix for each function tier.  These matrices show the accuracy of a function being correctly classified and, when incorrectly classified, which functions are selected instead of the true function. The color axis determines the occurrence ratio of the function classification. Ideally, high classification occurrence should be observed in matching indices (i.e., denser diagonal), indicating correct classification. As an example, we can observe that the model sometimes confuses the ``decrement'' class with ``increment'' and ``transmit'' classes in tier 3 function predictions.

\subsection{Feature Importance}
To investigate the contribution of the node and edge features to algorithm performance, we systemically drop features and observe the changes in F${_1}$-scores. Specifically, in this analysis, we drop single-node features, look at eliminating edge types, and lastly, remove all features from nodes and edges. By eliminating edge types and features, we look to discover if assembly edges and flow edges are more important toward prediction accuracy. In the edge importance analysis, we also look at retaining all edges without any features. We then eliminate all node and edge features to determine if graph topology impacts function predictions. Table \ref{table:featuretier1} shows the feature importance analysis per function tier. The results suggest that component basis has the highest impact on the performance among node features, whereas flow is the most influential edge feature. We also observe that initial node/edge features contribute more to the performance compared to topological information.

\subsection{Hierarchical Vs. Independent GNNs}
We investigate the contribution of introducing hierarchy on performance by comparing our hierarchical GNN framework with independently trained GNNs (i.e., no input from previous predictions). The results in Table \ref{table:hierarchical} suggest that introducing hierarchical training significantly improves the performance on tier 3, in which we observe an absolute 0.1 increase in micro F$_1$-score. We also see a high enhancement in tier 2 predictions. Because tier 1 GNNs do not have any predecessors in the hierarchy, they produce almost identical results in both cases. 

\begin{table*}[h!]
\setlength{\tabcolsep}{3.2pt}
\caption{\uppercase{\footnotesize Mean and Standard Deviation of precision, recall, and F$_1$-score on test set after 100 runs}}
\label{table:result}
\begin{center}
\begin{footnotesize}
\begin{sc}
\begin{tabular}{cl|ccc|ccc|ccc}
\toprule
\multicolumn{2}{c}{\textbf{Method}} & \multicolumn{3}{c}{\textbf{Micro}} & \multicolumn{3}{c}{\textbf{Macro}} & \multicolumn{3}{c}{\textbf{Weighted}} \\
\cmidrule{3-11}
& & Precision & Recall & F$_1$ & Precision & Recall & F$_1$ & Precision & Recall & F$_1$ \\
\midrule
\multirow{6}{*}{\begin{turn}{90}\textbf{Tier 1}\end{turn}}  
& Linear & 0.462 $\pm$ 0.02 & 0.462 $\pm$ 0.02 & 0.462 $\pm$ 0.02 & 0.599 $\pm$ 0.03 & 0.371 $\pm$ 0.01 & 0.390 $\pm$ 0.02 & 0.561 $\pm$ 0.02 & 0.462 $\pm$ 0.02 & 0.448 $\pm$ 0.02 \\
& MLP & 0.544 $\pm$ 0.02 & 0.544 $\pm$ 0.02 & 0.544 $\pm$ 0.02 & 0.706 $\pm$ 0.03 & 0.439 $\pm$ 0.02 & 0.479 $\pm$ 0.02 & 0.642 $\pm$ 0.02 & 0.544 $\pm$ 0.02 & 0.534 $\pm$ 0.02 \\
& SAGE \cite{hamilton_2017_nips}                     				& \textbf{0.832 $\pm$ 0.03} & \textbf{0.832 $\pm$ 0.03}  & \textbf{0.832 $\pm$ 0.03} & \textbf{0.882 $\pm$  0.03} & \textbf{0.712$\pm$ 0.04} & \textbf{0.773$\pm$ 0.04} & \textbf{0.846$\pm$ 0.03} & \textbf{0.832 $\pm$ 0.03} & \textbf{0.829$\pm$ 0.03} \\
& GCN \cite{kipf_2017_iclr}                      							& 0.795 $\pm$ 0.04 & 0.795 $\pm$ 0.04  & 0.795 $\pm$ 0.04 & 0.858 $\pm$ 0.02 & 0.662 $\pm$ 0.04 & 0.725 $\pm$ 0.04 & 0.817 $\pm$ 0.03 & 0.795 $\pm$ 0.04 & 0.791 $\pm$ 0.04  \\
& GAT \cite{velickovic_2018_iclr}            							& 0.794 $\pm$ 0.04 & 0.794 $\pm$ 0.04  & 0.794 $\pm$ 0.04 & 0.861 $\pm$ 0.03 & 0.668 $\pm$ 0.05 & 0.730 $\pm$ 0.05 & 0.818 $\pm$ 0.04 & 0.794 $\pm$ 0.04& 0.791 $\pm$ 0.04 \\
& GIN \cite{xu_2019_iclr}                         							 & 0.818 $\pm$ 0.04 & 0.818 $\pm$ 0.04  & 0.818 $\pm$ 0.04 & 0.877 $\pm$ 0.03 & 0.704 $\pm$ 0.04 & 0.766 $\pm$ 0.04 & 0.836 $\pm$ 0.03 & 0.818 $\pm$ 0.04 & 0.816 $\pm$ 0.04 \\
\midrule
\multirow{6}{*}{\begin{turn}{90}\textbf{Tier 2}\end{turn}} 
& Linear & 0.368 $\pm$ 0.01 & 0.368 $\pm$ 0.01 & 0.368 $\pm$ 0.01 & 0.522 $\pm$ 0.03 & 0.246 $\pm$ 0.01 & 0.268 $\pm$ 0.01 & 0.515 $\pm$ 0.02 & 0.368 $\pm$ 0.01 & 0.366 $\pm$ 0.02 \\
& MLP &  0.442 $\pm$ 0.02 & 0.442 $\pm$ 0.02 & 0.442 $\pm$ 0.02 & 0.587 $\pm$ 0.03 & 0.314 $\pm$ 0.02 & 0.346 $\pm$ 0.02 & 0.576 $\pm$ 0.03 & 0.442 $\pm$ 0.02 & 0.444 $\pm$ 0.02 \\
& SAGE \cite{hamilton_2017_nips}                     				& 0.756 $\pm$ 0.04 & 0.756 $\pm$ 0.04 & 0.756 $\pm$ 0.04 & \textbf{0.803 $\pm$ 0.04} & 0.618 $\pm$ 0.05 & 0.670 $\pm$ 0.05 & 0.774 $\pm$ 0.04 & 0.756 $\pm$ 0.04 & 0.750 $\pm$ 0.04 \\
& GCN \cite{kipf_2017_iclr}                     							& 0.714 $\pm$ 0.05 & 0.714 $\pm$ 0.05 & 0.714 $\pm$ 0.05 & 0.768 $\pm$ 0.04 & 0.562 $\pm$ 0.05 & 0.613 $\pm$ 0.05  & 0.740 $\pm$ 0.04 & 0.714 $\pm$ 0.05& 0.705 $\pm$ 0.05 \\
& GAT \cite{velickovic_2018_iclr}            							& 0.718 $\pm$ 0.06 & 0.718 $\pm$ 0.06 & 0.718 $\pm$ 0.06 & 0.785 $\pm$ 0.05 & 0.574 $\pm$ 0.07 & 0.627 $\pm$ 0.07 & 0.746 $\pm$ 0.05 &0.718 $\pm$ 0.06 & 0.710 $\pm$ 0.06\\
& GIN \cite{xu_2019_iclr}                          						 & \textbf{0.757 $\pm$ 0.04} & \textbf{0.757 $\pm$ 0.04} & \textbf{0.757 $\pm$ 0.04} & 0.802 $\pm$ 0.04 & \textbf{0.620 $\pm$ 0.05} & \textbf{ 0.672 $\pm$ 0.05} & \textbf{0.777 $\pm$ 0.04} & \textbf{0.757 $\pm$ 0.04} & \textbf{0.752 $\pm$ 0.04}\\
\midrule
\multirow{6}{*}{\begin{turn}{90}\textbf{Tier 3}\end{turn}} 
& Linear & 0.778 $\pm$ 0.07 & 0.778 $\pm$ 0.07 & 0.778 $\pm$ 0.07 & 0.776 $\pm$ 0.10 & 0.695 $\pm$ 0.10 & 0.699 $\pm$ 0.10 & 0.842 $\pm$ 0.05 & 0.778 $\pm$ 0.07 & 0.773 $\pm$ 0.08\\
& MLP & 0.779 $\pm$ 0.07 & 0.779 $\pm$ 0.07 & 0.779 $\pm$ 0.07 & 0.799 $\pm$ 0.10 &0.699 $\pm$ 0.10 & 0.714 $\pm$ 0.10 & 0.839 $\pm$ 0.06 & 0.779 $\pm$ 0.07 & 0.775 $\pm$ 0.07 \\
& SAGE \cite{hamilton_2017_nips}                      					&  0.783 $\pm$ 0.08 & 0.783 $\pm$ 0.08 & 0.783 $\pm$ 0.08 & 0.804 $\pm$ 0.10 & 0.711 $\pm$ 0.09 & 0.721 $\pm$ 0.09 & 0.844 $\pm$ 0.06 & 0.783 $\pm$ 0.08 & 0.777 $\pm$ 0.08\\
& GCN \cite{kipf_2017_iclr}                      							 & \textbf{0.787 $\pm$ 0.09} & \textbf{0.787 $\pm$ 0.09}  & \textbf{0.787 $\pm$ 0.09} & \textbf{0.831 $\pm$ 0.09} & \textbf{0.742 $\pm$ 0.10} &  \textbf{0.749 $\pm$ 0.10 }& \textbf{0.860 $\pm$ 0.07} & \textbf{0.787$\pm$ 0.09} & \textbf{0.786$\pm$ 0.09}\\
& GAT \cite{velickovic_2018_iclr}            							 & 0.779 $\pm$ 0.09 & 0.779 $\pm$ 0.09 & 0.779 $\pm$ 0.09 & 0.817 $\pm$ 0.09 & 0.719 $\pm$ 0.12 & 0.728 $\pm$ 0.11 & 0.844 $\pm$ 0.07 & 0.779 $\pm$ 0.09 & 0.775 $\pm$ 0.09 \\
& GIN \cite{xu_2019_iclr}                          							 &  0.758 $\pm$ 0.10 & 0.758 $\pm$ 0.10 &  0.758 $\pm$ 0.10 & 0.818 $\pm$ 0.10 &  0.724 $\pm$ 0.11 & 0.730 $\pm$ 0.11  & 0.842 $\pm$ 0.07 & 0.758 $\pm$ 0.10 & 0.756 $\pm$ 0.10\\
\bottomrule
\end{tabular}
\end{sc}
\end{footnotesize}
\end{center}
\end{table*}
\begin{table*}[h!] \vspace{-0.7\baselineskip}
\setlength{\tabcolsep}{2.5pt}
\caption{\uppercase{\footnotesize Feature importance for function prediction, when using GraphSAGE}}
\label{table:featuretier1}
\begin{center}
\begin{footnotesize}
\begin{sc}
\begin{tabular}{ll|ccc|ccc|ccc}
\toprule
\multicolumn{2}{c}{\textbf{Features}} & \multicolumn{3}{c}{\textbf{Tier 1 F$_1$-score}} & \multicolumn{3}{c}{\textbf{Tier 2 F$_1$-score}} & \multicolumn{3}{c}{\textbf{Tier 3 F$_1$-score}}\\
\midrule
Node & Edge & Micro & Macro & Weighted & Micro & Macro & Weighted & Micro & Macro & Weighted\\
\midrule
Com. Basis & All & \textbf{0.825 $\pm$ 0.03} & \textbf{0.770$\pm$  0.03} & \textbf{0.822$\pm$  0.03}  & \textbf{0.749$\pm$  0.04} & \textbf{0.670$\pm$  0.05} & \textbf{0.743 $\pm$  0.04} & \textbf{0.710 $\pm$ 0.10}  & \textbf{0.636$\pm$  0.11} & \textbf{0.700$\pm$  0.10}\\
Sys. Name & All &  0.639 $\pm$ 0.05  & 0.586 $\pm$ 0.05 & 0.632$\pm$ 0.05 &  0.555$\pm$  0.06  & 0.495$\pm$  0.07 &0.548 $\pm$ 0.06 & 0.549$\pm$  0.12 & 0.546 $\pm$ 0.12 &  0.550$\pm$  0.12\\
Sys. Type & All &  0.400 $\pm$ 0.08  & 0.339$\pm$  0.07 & 0.436$\pm$  0.09 & 0.278$\pm$  0.06  & 0.239$\pm$  0.06 & 0.309$\pm$  0.07 & 0.323$\pm$  0.12  & 0.323$\pm$  0.09 & 0.321$\pm$  0.12\\
Material & All & 0.616$\pm$  0.05 & 0.570$\pm$  0.05 & 0.608$\pm$  0.05 & 0.521$\pm$  0.05  & 0.469$\pm$  0.06 & 0.516 $\pm$ 0.05 &  0.443$\pm$  0.14 & 0.438$\pm$  0.10 & 0.416$\pm$  0.14\\
None & All & 0.585 $\pm$ 0.08 & 0.552$\pm$  0.09 &  0.588 $\pm$ 0.07 & 0.497 $\pm$ 0.08   & 0.452$\pm$  0.09 & 0.501$\pm$  0.08 & 0.513$\pm$  0.13  & 0.438 $\pm$  0.11 & 0.504$\pm$   0.12\\
\midrule
All & Flow & \textbf{0.877 $\pm$ 0.04} & \textbf{0.827 $\pm$ 0.04} & \textbf{0.876 $\pm$ 0.04} &  \textbf{0.824 $\pm$  0.05} & \textbf{0.753 $\pm$  0.05}  &\textbf{ 0.822 $\pm$  0.05} &  \textbf{0.919 $\pm$  0.09} & \textbf{0.844 $\pm$  0.14} & \textbf{0.918 $\pm$  0.10}\\	
All & Assem. & 0.535$\pm$  0.02  & 0.470 $\pm$ 0.02 & 0.527$\pm$  0.02 & 0.414$\pm$  0.03  & 0.314$\pm$  0.03 & 0.416 $\pm$ 0.03 & 0.689$\pm$  0.08  & 0.619$\pm$   0.11 & 0.696 $\pm$  0.08\\
All & None & 0.692$\pm$  0.06  & 0.614$\pm$  0.06 & 0.683 $\pm$ 0.06 &   0.613 $\pm$ 0.07   & 0.510$\pm$  0.07 &  0.604$\pm$  0.07 & 0.740$\pm$  0.08  & 0.657 $\pm$  0.09 &  0.736$\pm$   0.08\\
\midrule
All & All & 0.832 $\pm$ 0.03 & 0.773$\pm$ 0.04 &0.829$\pm$ 0.03 & 0.756 $\pm$ 0.04 & 0.670 $\pm$ 0.05 & 0.750 $\pm$ 0.04 & 0.783 $\pm$ 0.08 &0.721 $\pm$ 0.09 & 0.777 $\pm$ 0.08\\
None & None & 0.147 $\pm$ 0.04  &  0.129$\pm$  0.03 & 0.137$\pm$  0.04  & 0.068$\pm$  0.04 &  0.045$\pm$  0.03 & 0.070 $\pm$ 0.04 & 0.163 $\pm$  0.09 & 0.128$\pm$   0.07 & 0.194$\pm$   0.10\\
\bottomrule
\end{tabular}
\end{sc}
\end{footnotesize}
\end{center}
\vskip -0.1in
\end{table*}
\begin{figure*}
    \centering
    \subfigure[]{\includegraphics[width=0.33\textwidth]{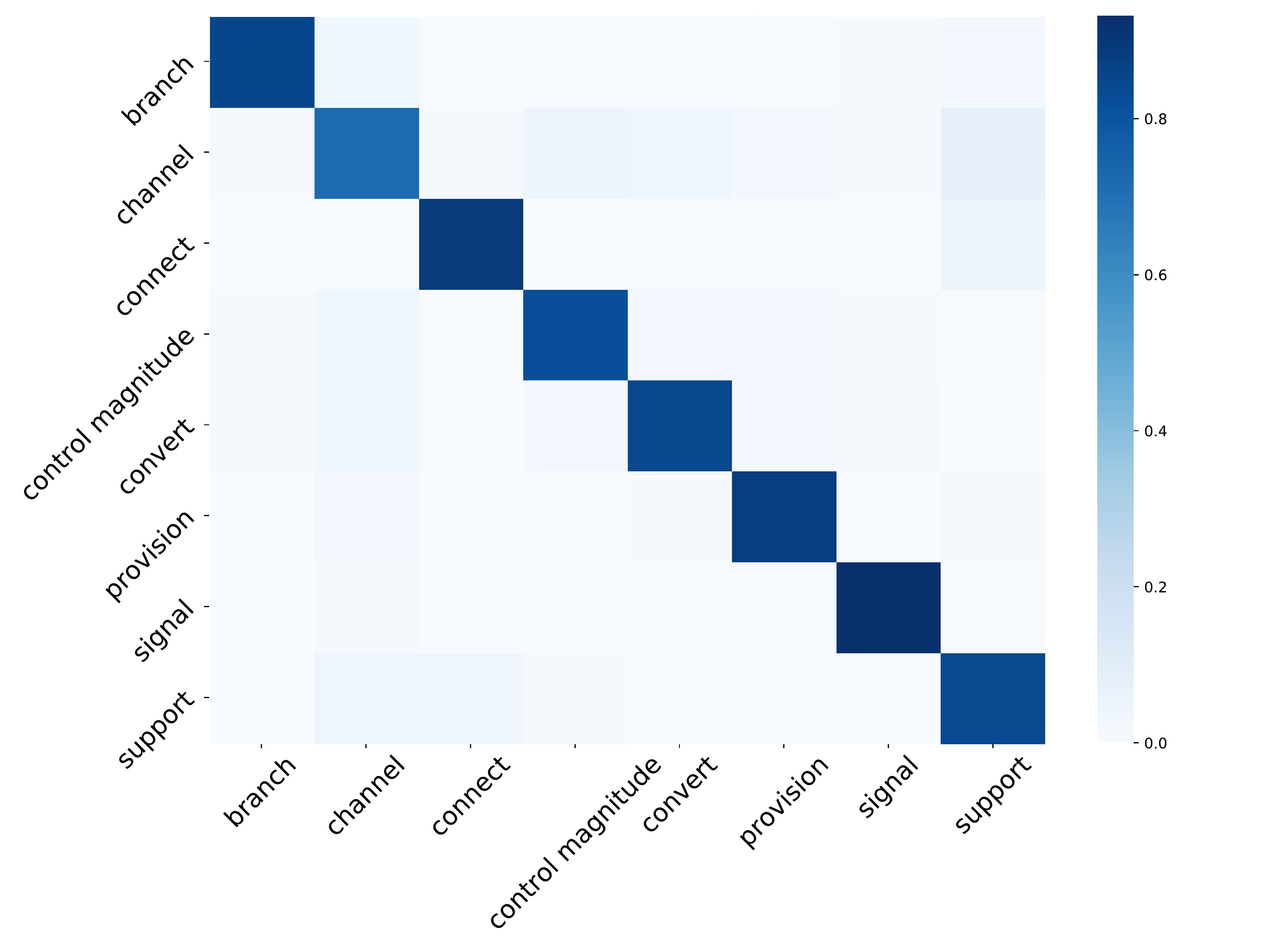}} 
    \subfigure[]{\includegraphics[width=0.33\textwidth]{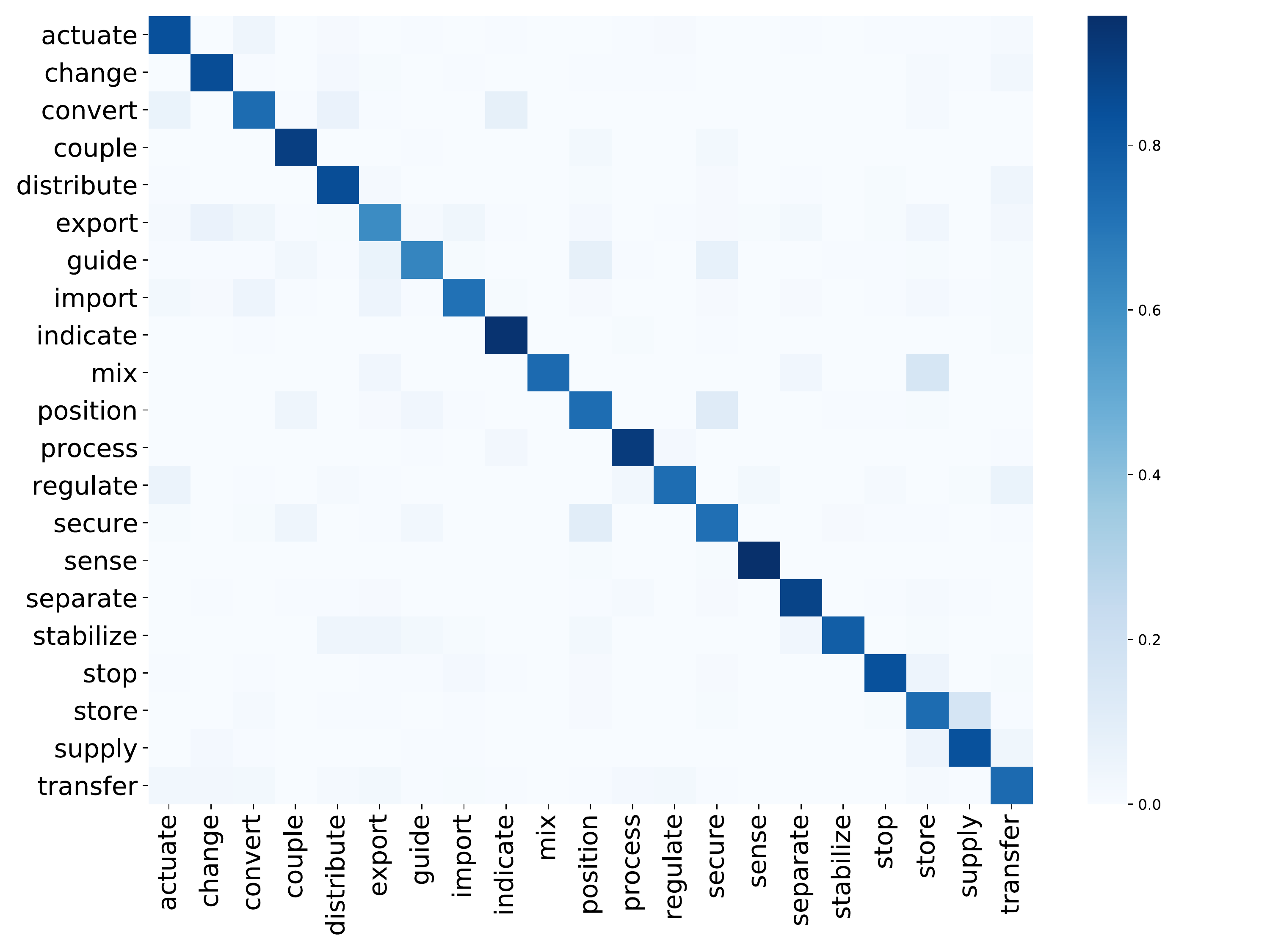}} 
    \subfigure[]{\includegraphics[width=0.33\textwidth]{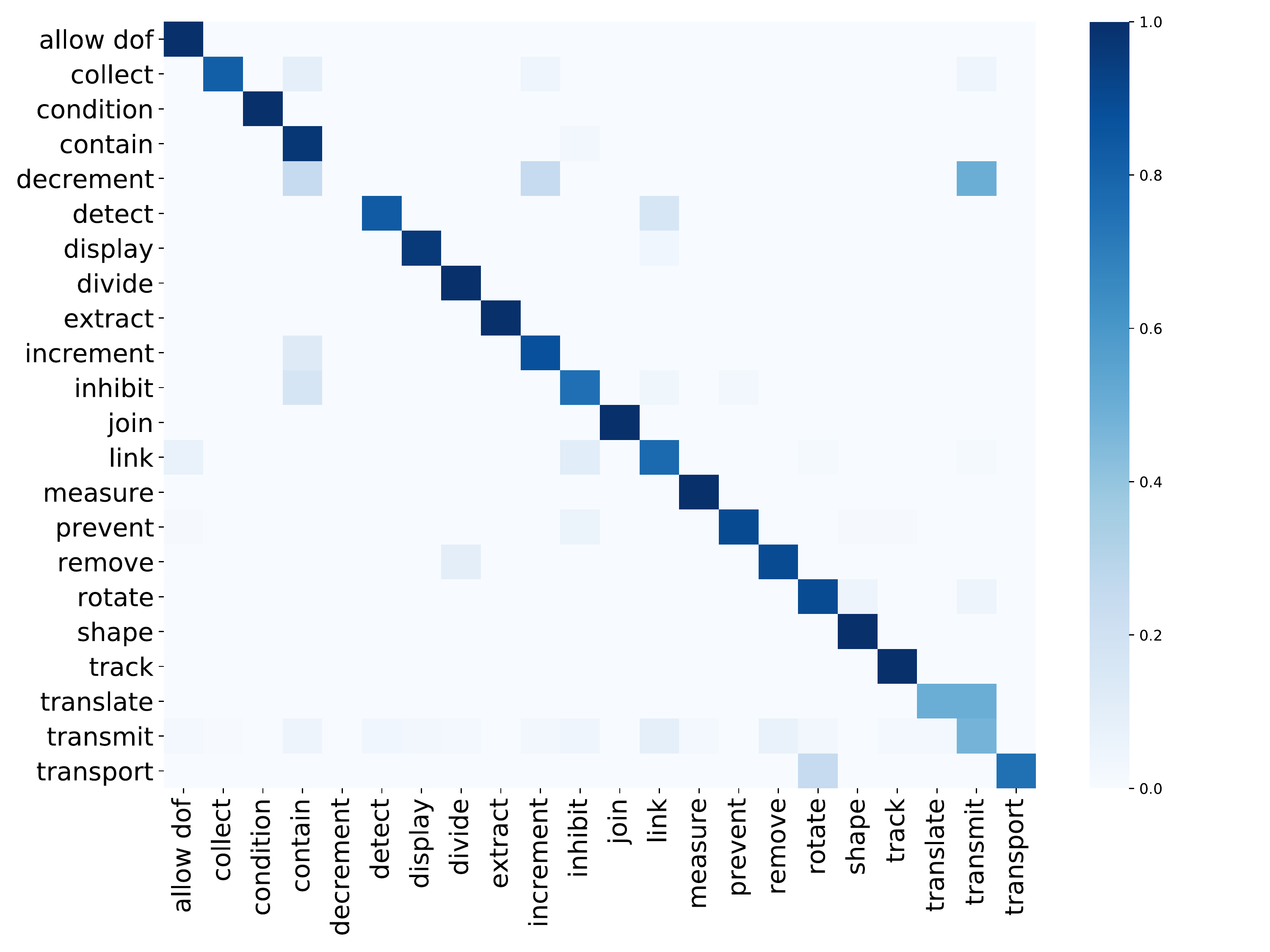}\label{fig:confusion3}}
    \caption{\uppercase{\footnotesize Confusion matrices of (a) tier 1, (b) tier 2, and (c) tier 3 function predictions. The rows represent the ground truth whereas the columns represent the predictions}}
    \label{fig:confusion}
\end{figure*}
\begin{table}[h!] \vspace{-0.7\baselineskip}
\setlength{\tabcolsep}{2.5pt}
\caption{\uppercase{\footnotesize Performance of hierarchical and independent GNNs}}
\label{table:hierarchical}
\begin{center}
\begin{footnotesize}
\begin{sc}
\begin{tabular}{llccc}
\toprule
\textbf{Tier} &\textbf{Method} & \multicolumn{3}{c}{\textbf{F$_1$-score}}\\
\cmidrule{3-5}
~ & ~ & Micro & Macro & Weighted \\
\midrule
\multirow{2}{*}{\textbf{1}} 
& Hierarchical & 0.832 $\pm$ 0.03 & 0.773 $\pm$ 0.04 & 0.829 $\pm$ 0.03 \\
& Independent & 0.834 $\pm$ 0.03 & 0.771 $\pm$ 0.04 & 0.827 $\pm$ 0.03 \\
\midrule
\multirow{2}{*}{\textbf{2}} 
& Hierarchical &\textbf{ 0.756 $\pm$ 0.04} & \textbf{0.670 $\pm$ 0.05} & \textbf{0.750 $\pm$ 0.04} \\
& Independent &  0.713 $\pm$ 0.05 & 0.644 $\pm$ 0.04 & 0.719 $\pm$ 0.05 \\
\midrule
\multirow{2}{*}{\textbf{3}}
& Hierarchical & \textbf{0.783 $\pm$ 0.08} &\textbf{0.721 $\pm$ 0.09} & \textbf{0.777 $\pm$ 0.08}\\
& Independent & 0.684 $\pm$ 0.09  &  0.656 $\pm$ 0.06  &  0.683 $\pm$ 0.07 \\
\bottomrule
\end{tabular}
\end{sc}
\end{footnotesize}
\end{center}
\vskip -0.1in
\end{table}

\subsection{Assumptions and Limitation}
The OSDR is a multi-decade-long project that has been manually influenced by many organizations and design engineers. Knowing this, we reiterate that the data from the repository is unbalanced, sparse, and often non-congruent. We observe noted cascading label imbalance and absence from higher-order hierarchy taxonomy to lower-order taxonomy terms. Tier 3 classifications in function, flow, and component basis terms per component are often missing assignment. In short, the OSDR shows a user-curated bias toward defining higher-order basis terms in both function and flow, as shown in Figures \ref{fig:frequency} and \ref{gnna}. Although the data is incomplete and imbalanced, we maintain that the compilation of the OSDR is representative of knowledge from many design engineers with varying expertise. As such, the OSDR can be thought of as a sampling of function-based domain knowledge ranging from novice to expert design engineers.

In using a hierarchical GNN model, we inherit the assumptions that were used to create and facilitate the propagation of such taxonomies. Function, flow, and component taxonomies are directional, but parent-child assembly relationships are not. As such, in our graph representations, flow edges are directional, whereas assembly edges are not. The GNN model takes into account direction information and models non-directional assembly edges as bi-directional. We recognize that the OSDR taxonomy approach is one of many adopted function, flow, and component standardizations. These taxonomies are applied to a wide breadth of consumer products. We choose to adopt the OSDR taxonomies as a starting point, but we realize that this definition of function might be generalizable to all design problems and domains.

\section{DISCUSSION}
As shown in section \ref{results}, we observe that the overall performance of the GNN with GraphSAGE layers is strong, particularly in tier 1 function prediction. Tier 2 and 3 function predictions are competitive against other GNN types, only coming second to GNN with GIN (tier 2) and GNN with GCN (tier 3) layers. These results exceeded the general expectation of all GNN types, given the unbalanced and very sparse product design data. Given a repository of just 160 products spanning various industries (automotive, consumer goods, furniture), it is encouraging that the proposed GNN architecture was able to ascertain part-level functional classification with a micro-average F${_1}$-score of 0.790. We incorrectly anticipated that function would be more product family-specific and would cause model confusion between industry domains. Future work could look at improving performance by collecting more data, and including more complex products, possibly from different industries.

\textit{Where} the model fails can be identified through the relative performance between function classes. The confusion matrices shown in Figure \ref{fig:confusion} could serve as a valuable tool for a practitioner wanting to adopt our method, as they explicitly show the relative performance of all classes of function we can predict. The confusion matrices also suggest cascading false negatives and false positives as our model moves from tier 1 through tier 3 function predictions. We theorize this is caused by the significant scarcity of tier 3 function data, as shown in Figure \ref{fig:frequency}. Moreover, tier 2 and 3 suffer from more significant data imbalance in comparison to tier 1. In context, the concatenation of a high number of classification labels and data imbalance found in tier 2 and 3 functions resulted in some meaningful false negatives and false positives during testing. 

\textit{Why} the model fails could be attributed to subjectivity in the function definitions and to data imbalance caused by the overall OSDR embedded bias toward defining tier 1 functions and solid flows. In the tier 3 predictions shown in Figure \ref{fig:confusion}, we observe that the model often confuses function labels that are related. For example, ``decrement'' is often confused with ``increment'' and ``transmit". In the same regard, ``translate" is often confused with ``transmit". The model appears to ascertain the contextual function correctly but has trouble discerning the details that individualize some tier 3 functions. In this example, the GNN model finds that these unlabeled components generally are ``moving" flow. However, the model can not classify if the component is ``incrementing", ``decrementing", or translating a material flow or ``transmitting" a signal or energy flow. 
Confusion in low-frequency function classes can be also be attributed to conflicting knowledge caused by sparse edges, especially considering confusion between material flows and the other flows. 
Moreover, the results in Table \ref{table:result} and Table \ref{table:featuretier1} show that macro scores are usually lower than micro scores, indicating that the least populated classes are poorly classified relative to the more populated classes. Based on this, further application of this work should collect additional data for the least populated classes. Researchers looking to apply these methods would benefit by augmenting the current dataset to address data imbalance and scarcity guided by Figure \ref{fig:frequency}, while also modifying the dictionary of functions to suit their task.

Table \ref{table:featuretier1} shows an adversarial effect between flow edges and assembly edges. When only considering flow edges, the GNN model performed better than with both edge types. Conversely, when just considering assembly edges, performance sharply declines. Upon discovering this effect, we theorized that energy and signal flows are not always correlated with physical assembly or ``closeness" of components that are inputting or outputting these types of flows. While there is significant overlap between the two edge types, the slight differences in flow and assembly edges are enough to cause the adversarial effect. When we highlight only solid or material flows with assembly flows, we find that there is a significant increase in F${_1}$-score (\textbf{0.754}, \textbf{0.670}, \textbf{0.907} micro-average) that begins to recapture the performance of the only flow GNN model output (\textbf{0.887}, \textbf{0.824}, \textbf{0.919} micro-average). The inclusion of only material flow and assembly edges is more precise in tier 3 function classification than the all flow and assembly edges model (\textbf{0.832}, \textbf{0.756}, \textbf{0.783} micro-average).  Moving forward, this finding is advantageous in future work considering geometric and CAD embeddings. Whereas in CAD data, it might be challenging to capture energy and signal flows, it might be more promising to capture solid flows. As such, solid flows are likely the most analogous bridge between assembly and function. 

To evaluate the success of our method, it is helpful to look at the top-k predictions by the GNN, as this would be more analogous to how the method would be applied in a use case. By selecting the correct result from the top-3 predictions we achieve significant improvement in F${_1}$-score (\textbf{0.985}, \textbf{0.914}, \textbf{0.935} micro-average) over the top-1 F${_1}$-scores reported in Table \ref{table:result}. This shows that future work could benefit from implementing a temperature-based probability sampling approach from the top-3 predictions to further improve the quality of the method. In addition, future work should look at establishing a comparison between the method described in this work and a human baseline, as this would also provide insights into the biases of the current dataset and ambiguities around function definition.

\section{CONCLUSION}
In this work, we use graph neural networks to classify the function of parts in an assembly given design knowledge about the part, such as the semantic name, the material, the assembly connections, and the energy flowing into and out of the part. Here we extract data from 160 products in the OSDR and represent it within 160 graphs and a total of 15,636 nodes, with each node containing design knowledge about the part in a multi-dimensional feature vector. With this data, we are able to train a GNN to predict the function of a part with micro precision of 0.832 for tier 1 (broad), 0.756 for tier 2, and 0.783 for tier 3 (specific) functions. Our results suggest that the hierarchical structure of products and relevant design knowledge describing sub-components can be learned effectively with graph neural networks. The quality of these results show promise in supporting the development of a larger function dataset from a more extensive set of products. Our method could be further developed by learning from the geometric data of the part, a prominent design feature that is missing from the current work in lieu of the semantic name of the part.  

There are several research directions to expand on this work. By inferring the function of a design at any point in the design process, an intelligent design agent could better support the designer throughout various tasks, such as the design of complex systems like industrial machinery. The quality required from the predictive system will be dictated by the task being performed. For some use cases, high-quality top-3 predictions could be sufficient to give enough context to the design agent to support the designer. For example, function data could support the designer during the conceptual design stage in assessing the feasibility of a design \cite{berquand2019}, searching for functionally similar parts \cite{lupinetti2019}, or by enabling automated functional modeling \cite{Mikes2020a, Edmonds2020, cheng2017}. In the detail design stages, it could aid in verifying the satisfaction of higher-level design requirements \cite{erden2008}. Furthermore, this work could further the development of function-based sustainability methods and other function-related environmental considerations during the early design phases \cite{Arlitt2017,Devanathan2010,Bohm2010}. A human-centric case study should be conducted to establish a baseline against which the method presented in this work can be evaluated.

In future work, we look to enable knowledge-based CAD systems through automated function inference by bridging a gap in understanding between the designer and an intelligent design agent. We envision design tools extending beyond documentation, simulation, and optimization towards intelligent reasoning tasks that help designers make informed design decisions. 

\begin{acknowledgment}
This material is based upon work supported by the National Science Foundation under Grant No. CMMI-1826469. Any opinions, findings, and conclusions or recommendations expressed in this material are those of the author(s) and do not necessarily reflect the views of the National Science Foundation.
\end{acknowledgment}

%
% ...

\bibliographystyle{asmems4}

\bibliography{asme2e}


\appendix       
\begin{appendices}
\onecolumn
\section{Appendix A: Data Example from the Oregon State Design Repository}
\label{appendix:data}

\begin{table*}[h!]
\begin{center}
\begin{footnotesize}
\caption{\uppercase{Vegetable Peeler Example Product Data}}
\label{BOM}
\begin{tabular}{c l l l l l l l}
\hline
System & ID & Component & Child of & Material & Input Flow - From & Output Flow - To & Function Tier 1/2/3 \\
\hline
Vegetable Peeler & 1 & Unclassified & - &- &  - &- &- \\
\midrule
\midrule
-& 2 & Blade& 1& Steel & Solid - Int* & Solid - Int & Branch/Separate/- \\
-& 2& Blade& 1& Steel & Solid - Ext** & Solid - Int & Channel/Import/- \\
-& 2& Blade& 1& Steel & Solid - Int & Solid - Ext & Channel/Export/- \\
-& 2& Blade& 1& Steel & Solid - Int & Solid - Ext & Channel/Export/-  \\
-& 2& Blade& 1& Steel & Mechanical - 3 & Mechanical - Ext & Channel/Export/-\\
-& 2& Blade& 1& Steel & Solid - Int & Solid - Int & Channel/Guide/- \\
-& 2& Blade& 1& Steel & Status - Int & Status - Ext  & Signal/Indicate/- \\
-& 2& Blade& 1& Steel & Solid - 1 & Solid - int & Support/Secure/- \\
-& 3& Handle& 1& Plastic & Control - Ext & Control - Int & Channel/Import/-  \\
-& 3& Handle& 1& Plastic & Human - Ext & Human - Int & Channel/Import/- \\
-& 3& Handle& 1& Plastic & Human Energy - Ext & Human Energy - Int & Channel/Import/-  \\
-& 3& Handle& 1& Plastic & Human - Int & Human - Ext & Channel/Import/- \\
-& 3& Handle& 1& Plastic & Human Energy - Int & Mechanical - 2& Convert/-/-  \\
-& 3& Handle&1 & Plastic & Solid - 2 & Solid - Int & Support/Secure/-  \\
\hline
\end{tabular}
* Int (Internal) is nonspecific flows from inside the system \\
** Ext (External) is nonspecific flows from outside the system \\
\end{footnotesize}
\end{center}
\end{table*}

\newpage

\section{Appendix B: Statistics}

\begin{table*}[h!] \vspace{-0.7\baselineskip}
\setlength{\tabcolsep}{2.5pt}
\caption{\uppercase{\footnotesize Statistics of graphs}}
\label{table:graph_stats}
\begin{center}
\begin{footnotesize}
\begin{sc}
\begin{tabular}{l|cccc|ccc|cc}
\toprule
~ & Mean & STD & Min & Max & 0.25 Quantile & 0.5 Quantile & 0.75 Quantile & Skewness & Kurtosis \\
\midrule

\#nodes & 97.72, & 100.01 & 3.00 & 930.00 & 42.50 & 79.50 & 125.50 & 4.62 & 31.51 \\
\#edges & 790.71 & 1039.16 & 0.00 & 9634.00 & 180.50 & 461.50 & 981.25 & 4.45 & 31.42 \\
density & 0.11 & 0.13 & 0.00 & 1.50 & 0.06 & 0.08 & 0.13 & 7.51 & 73.52 \\
degree & 13.29 & 7.82 & 0.00 & 58.73 & 8.44 & 11.13 & 17.10 & 2.33 & 9.61 \\
\bottomrule
\end{tabular}
\end{sc}
\end{footnotesize}
\end{center}
\vskip -0.1in
\end{table*}

\begin{itemize}
    \item 30.82\% of edges are assembly and 69.18\% are flows.
    \item 3 out of 160 graphs are DAG.
\end{itemize}

\begin{figure*}[h!]
    \centering
    \includegraphics[width=170mm]{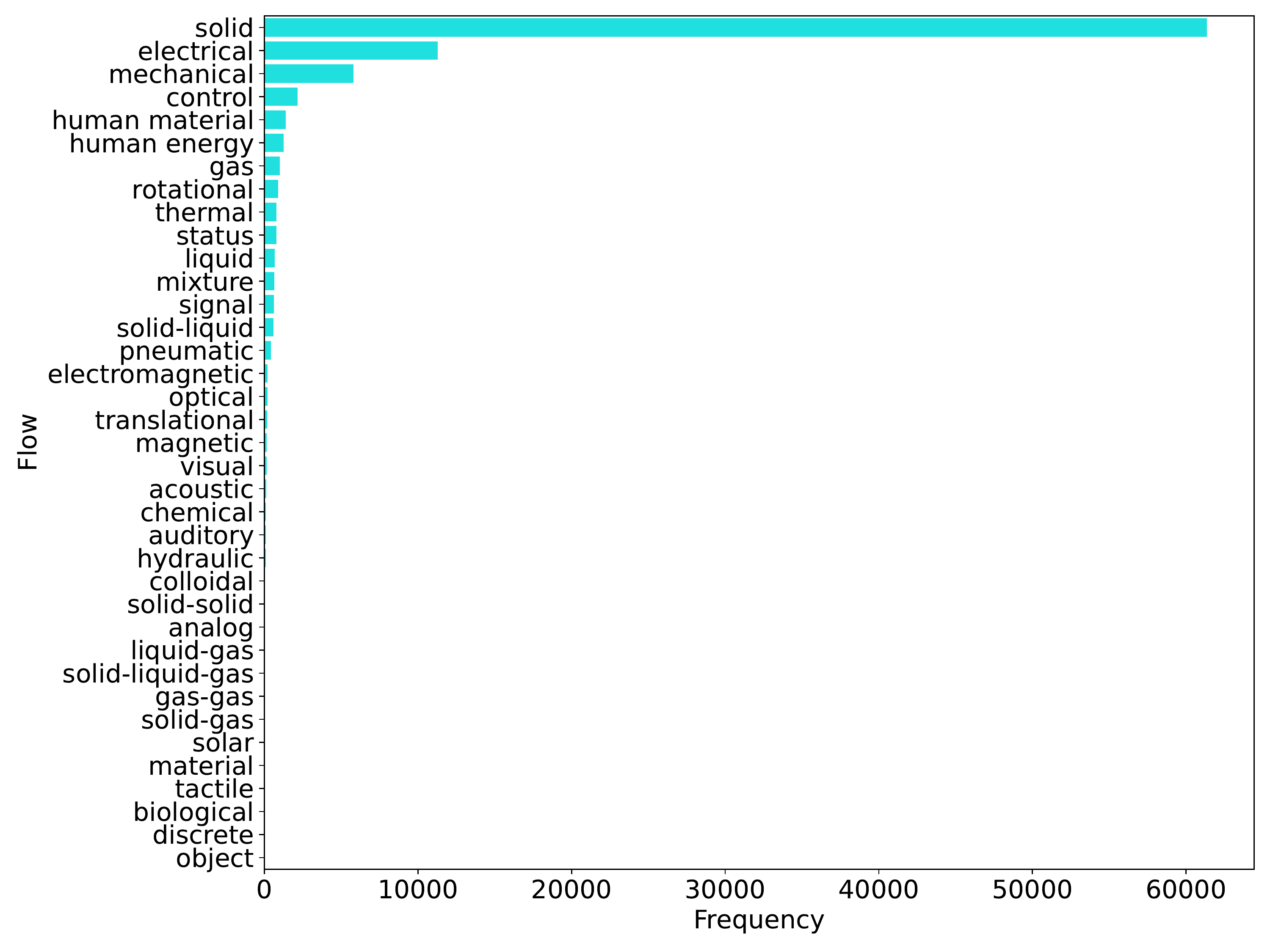}
    \caption{\uppercase{\footnotesize Distribution of flow edges}}
    \label{gnna}
\end{figure*}

\section{Appendix C: Hyperparameters \& Architecture}
For GNNs, we represent the edge features as follows. We represent the in-flow and out-flow features as one-hot representations and the assembly link as a single indicator. Thus, if there are N unique flows in the dataset, the initial edge features are represented as a $2N+1$ vector. We concatenate the one-hot representations of the node features to create the initial node feature. For the linear and MLP baselines, we first project the initial node and features to embeddings of the same size using two dedicated linear layers. We then represent each node by summing up its embedding with the summation of the dot product of all the neighbor node-edge pairs. We then pass the computed embeddings to MLP or linear models. As mentioned, we choose the number of GNN layers and hidden dimension size from the range of [1, 2, 3] and [64, 128, 256], respectively. We found that for GraphSage and GCN, two layers with a dimension of 128 yield the best results, whereas for GIN a network with three layers and 256 dimensions, and for GAT, a network of one layer and 128 dimensions, results in the best performance.

\end{appendices}

\end{document}